\documentclass[runningheads]{llncs}
\usepackage{graphicx}

\usepackage{tikz}
\usepackage{comment}
\usepackage{amsmath,amssymb} 
\usepackage{color}

\usepackage{booktabs}
\usepackage{xcolor}
\usepackage{multirow}
\usepackage{enumitem}
\usepackage{wrapfig}
\usepackage{caption}
\usepackage{flushend}
\usepackage{xargs}

\usepackage{xspace}
\newcommand{\AC}{ComplETR\xspace}

\usepackage[accsupp]{axessibility}  

\begin{document}
\pagestyle{headings}
\mainmatter
\def\ECCVSubNumber{}  

\title{\AC: Reducing the cost of annotations for object detection in dense scenes with vision transformers}

\titlerunning{\AC}
%
\author{Achin Jain \inst{1}\thanks{Corresponding author. $^\dagger$ Work done at AWS.} \and
Kibok Lee \inst{1,2}$^\dagger$ \and
Gurumurthy Swaminathan \inst{1} \and
Hao Yang \inst{1} \and
Bernt Schiele  \inst{1} \and
Avinash Ravichandran \inst{1} \and
Onkar Dabeer\inst{1}}
\authorrunning{A.~Jain et al.}
%
\institute{$^1$ AWS AI Labs \ \ $^2$ Yonsei University 
\email{\{achij,gurumurs,haoyng,bschiel,ravinash,onkardab\}@amazon.com} \\
\email{kibok@yonsei.ac.kr}
}
\maketitle

\begin{abstract}
Annotating bounding boxes for object detection is expensive, time-consuming, and error-prone. In this work, we propose a DETR based framework called {\it \AC} that is designed to explicitly complete missing annotations in partially annotated dense scene datasets. This reduces the need to annotate every object instance in the scene thereby reducing annotation cost. {\it \AC} augments object queries in DETR decoder with patch information of objects in the image.
Combined with a matching loss, it can effectively find objects that are similar to the input patch and complete the missing annotations. We show that our framework outperforms the state-of-the-art methods such as Soft Sampling and Unbiased Teacher by itself, while at the same time can be used in conjunction with these methods to further improve their performance. Our framework is also agnostic to the choice of the downstream object detectors; we show performance improvement for several popular detectors such as Faster R-CNN, Cascade R-CNN, CenterNet2, and Deformable DETR on multiple dense scene datasets.
\keywords{Object detection, dense scenes, partial annotations}
\end{abstract}

\section{Introduction}

\begin{figure}[t!]
\small
\centering

\includegraphics[width=0.49\linewidth]{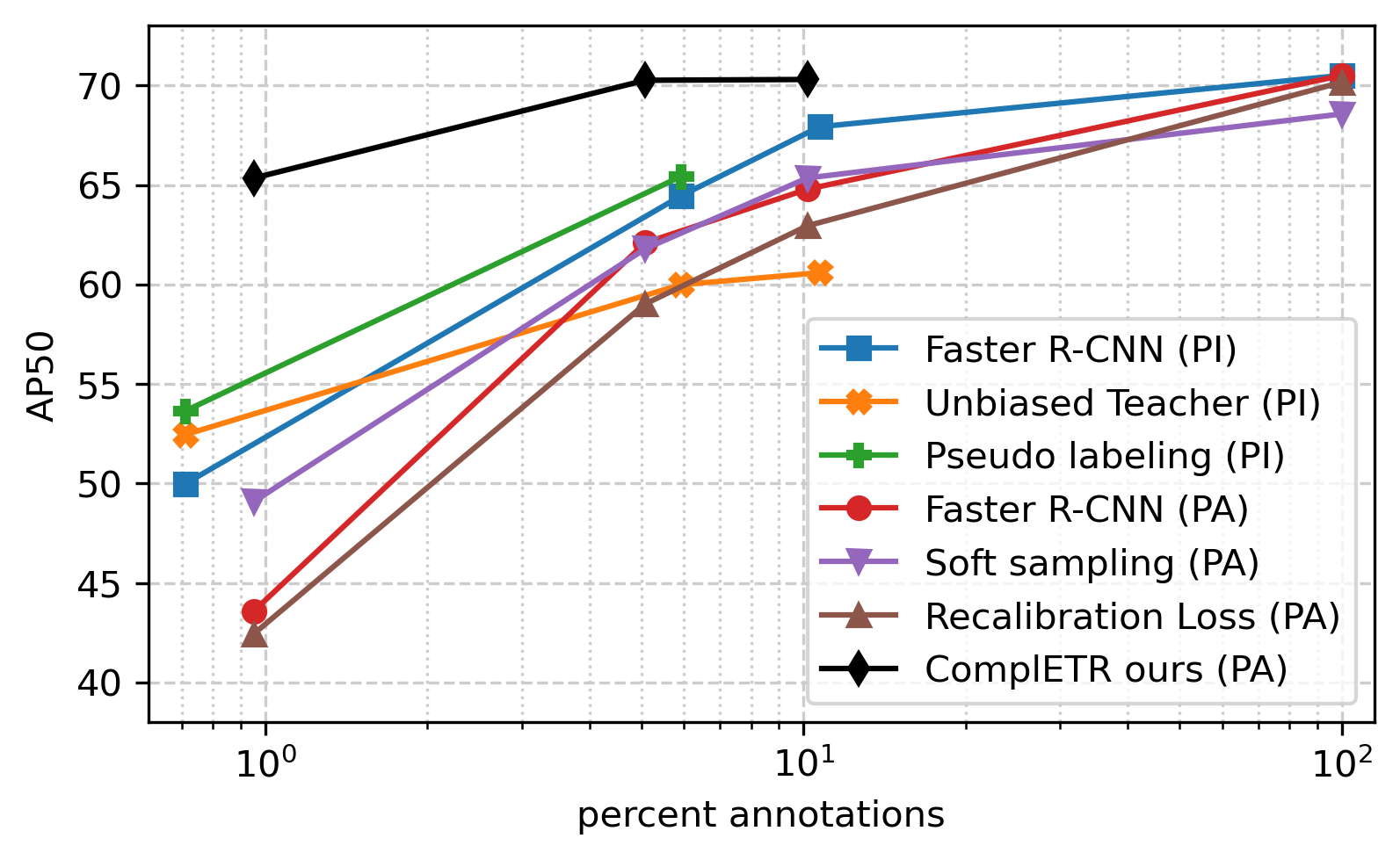}
\hfill
\includegraphics[width=0.24\linewidth]{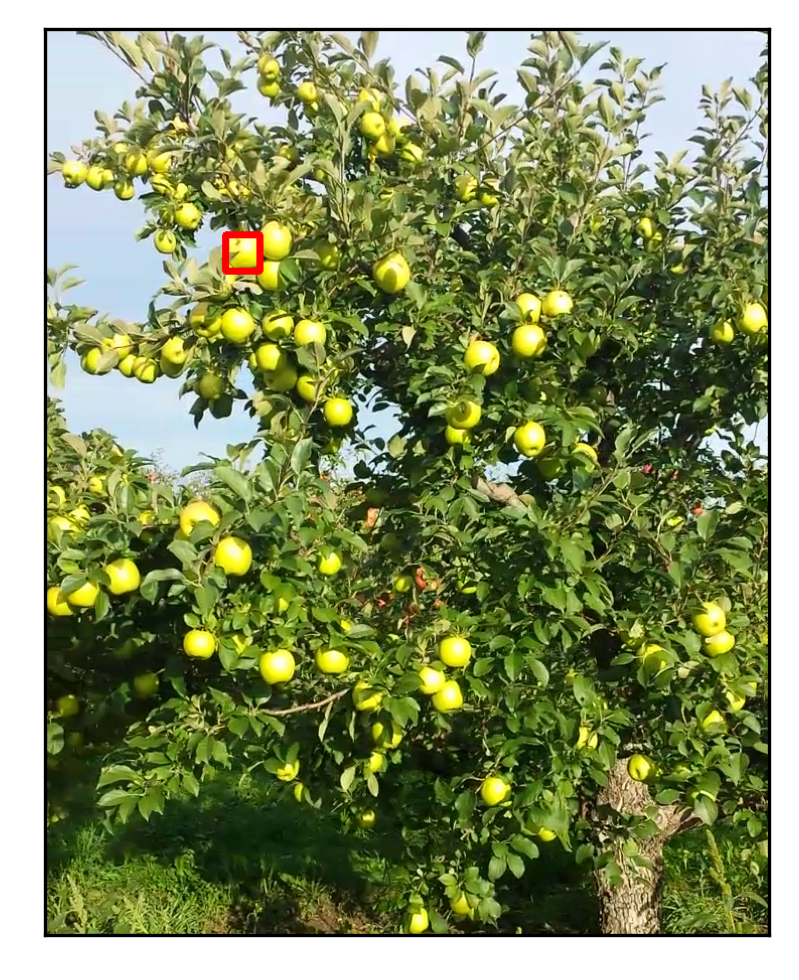}
\hfill
\includegraphics[width=0.24\linewidth]{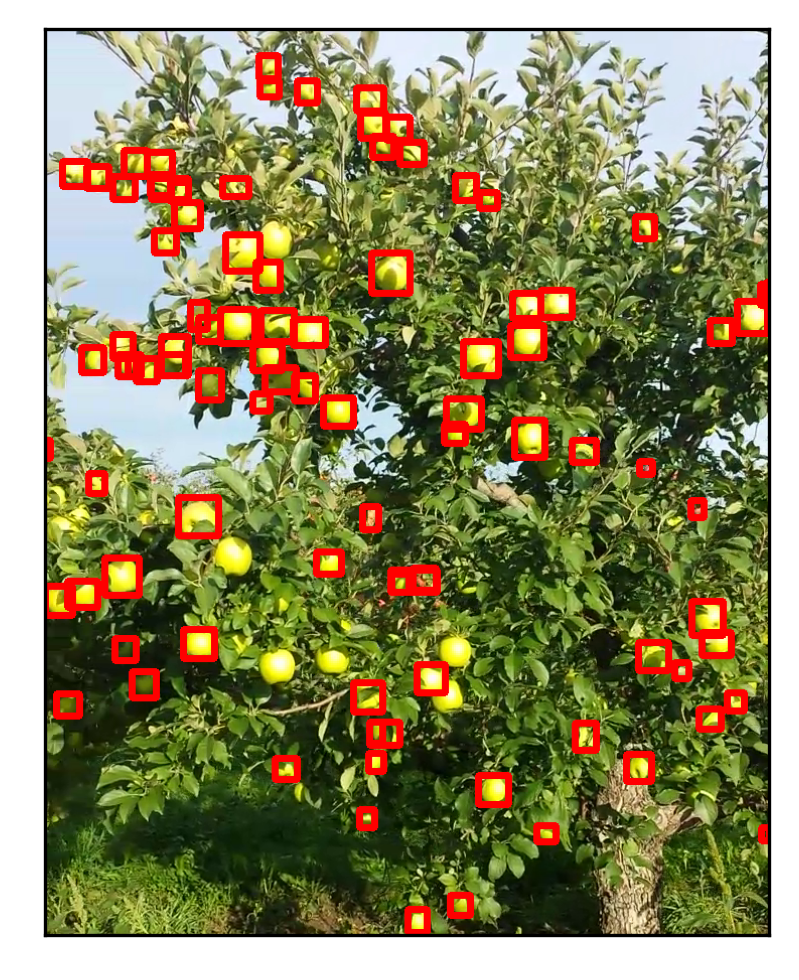}

\caption{Left: \AC outperforms all baselines with similar annotation budget on Minneapple~\cite{Minneapple} dataset. PI refers to the partial images setting and PA to the partial annotations setting. Middle: an example with partial annotation (only 1 out of more than 50 instances is annotated). Right: \AC predicts the missing annotations.}
\label{F:examples}
\vspace{-15pt}
\end{figure}

State-of-the-art object detection approaches require expensive and challenging bounding box annotations for training.
According to~\cite{hao2012crowdsourcing}, drawing a bounding box and verifying its quality takes about 35 seconds per object. This is time-consuming and particularly problematic for detection in dense scenes with tens or hundreds of objects in each image; examples include detecting bees for non-invasive tracking~\cite{Bees} (avg.~22 objects per image), humans in crowd~\cite{CrowdHuman} (avg.~22 objects per image), apples on trees~\cite{Minneapple} (avg.~48 objects per image; see Figure~\ref{F:examples}), vehicles or empty parking spots~\cite{CARPK} (avg.~59 objects per image), products in retail stores~\cite{SKU110K} (avg.~147 objects per image), etc.
For such datasets, human annotation cost exceeds several thousands of hours.

To alleviate the annotation cost, a common approach in the literature is to use semi-supervised learning (SSL), where all objects in a subset of images are \textit{fully} annotated, and other images are unannotated. We refer to this setting as \textit{partial images}. State-of-the-art methods in SSL have achieved great success in benchmark datasets such as COCO~\cite{COCO} or VOC~\cite{VOC}. In this paper, we argue that SSL is not the optimal way to reduce annotation cost, specifically for object detection in dense scenes where the average number of objects per image is large (e.g.~more than 20). This is because the current methods in SSL do not consider an important property in such datasets: visually similar objects appear many times within the same or across images in different perspective, size, color, etc. and hence annotating all objects that are similar may not add additional information. 

Alternatively, we can annotate some objects (leaving others unannotated) in all (or most) training images, which we refer to as \textit{partial annotations}. We believe this setting has three benefits:
(1) It is less tedious to annotate as it alleviates the need for intensive quality assurance to \textit{ensure that an image is fully annotated} which can increase the annotation cost. In contrast, asking an annotator to annotate $K$ objects per image is a much easier guideline to communicate and control the quality rather than asking them to annotate all objects in the image.
(2) Sampling bias can be an issue in partial images if we are not careful with the image selection; simple random sampling may lead to sub-optimal results.
(3) When combined with our proposed method to exploit inter- and intra- image similarities within class objects, we can perform significantly better than the state-of-the-art SSL methods used in the partial images setting.

Specifically, we introduce a framework called \textit{\AC} (\textit{Compl}eter with D\textit{ETR} style architecture) that is designed to explicitly complete annotations in partially annotated images.
The underlying principle of \AC is to utilize the intra-class similarity within and across images. Given several annotated objects in a training image, our proposed method explicitly mines and finds other visually similar objects within and across training images.
For example, given sparse annotations in the middle image in Figure~\ref{F:examples}, \AC utilizes the similarity between ``apple'' class to complete missing annotations on the right.

Naively, this matching process can be expensive and require significant network architecture modifications for conventional deep learning-based object detection methods. 
With the recent development of transformer-based architectures for objects detection such as DETR~\cite{carion2020end}, we can achieve this goal of finding similar objects by exploiting the cross-attention between object queries and encoder features in the transformer decoder. This enables us to train the system with similar cost and architecture to a standard object detector.

We also propose a decoupled offline training process for \AC. Specifically, it is first trained on partially annotated images and is then used to explicitly complete the missing annotations. Subsequently, any detector can be trained on the generated (pseudo) ground-truths. This decoupling provides us several practical advantages compared to an online teacher-student model.
Firstly, it makes \AC agnostic to the choice of an object detector, i.e., any off-the-shelf detector, regardless of its architecture, like Faster R-CNN\cite{ren2015faster}, DETR~\cite{carion2020end} or CenterNet2~\cite{duan2019centernet} can benefit from the completed annotations depending on the desired latency/accuracy trade-off.
In contrast, the widely used online teacher-student SSL methods need the teacher and student models to be exactly the same in order to update the teacher model with exponential moving average (EMA) of the student model. Even if it is possible to change the architecture from the commonly used Faster-RCNN to a different one (such as DETR) it may still need extensive hyper-parameters search for optimal performance.
Secondly, the completed annotations from \AC are interpretable and can be visually inspected with human-in-the-loop verification to further improve the quality of the completed annotations before training, if needed.

This paper makes the following contributions:
\vspace{-\topsep}
\begin{enumerate}[noitemsep,leftmargin=*]
    \item We show that the partial annotations setting (few annotations in all images) is a better alternative to the partial images setting (all annotations in few images) to alleviate annotation cost for \textit{object detection in dense scenes} (number of objects per image is large). We show that it can achieve significant performance improvement when combined with the proposed annotation completer.
    
    \item We propose \AC that explicitly completes missing annotations in the partial annotations setting thereby reducing the need for annotating every object. We show that our framework by itself not only outperforms the state-of-the-art methods like soft sampling~\cite{wu2019soft}, background recalibration loss~\cite{zhang2020solving} in partial annotations and Unbiased Teacher~\cite{liu2021unbiased} in the partial images setting with a similar annotation budget, but is also complimentary to them and can further improve their performance.
    
    \item We propose a decoupled training process for \AC which makes it agnostic to the choice of an object detector trained on the completed annotations. We show that it improves the performance of several popular detectors like Faster R-CNN~\cite{ren2015faster}, Cascade R-CNN~\cite{cai2018cascadercnn}, CenterNet2~\cite{duan2019centernet}, Deformable DETR~\cite{zhu2021deformable} with evaluation on five datasets from dense scenes: Minneapple~\cite{Minneapple}, Bees~\cite{Bees}, CrowdHuman~\cite{CrowdHuman}, LVIS~\cite{LVIS}, and COCO~\cite{COCO}.
\end{enumerate}

\section{Related work}
\label{S:related}

To reduce the amount of annotation effort for object detection, methods proposed in the literature range from
(1) \textit{weakly-supervised learning} that uses alternate forms of less expensive annotations, such as image-level labels, center points, and scribbles in the place of bounding boxes \cite{bilen2016weakly,gao2018notercnn,ren2020instance,ren2020ufo2},
(2) \textit{semi-supervised learning} (or partial images setting) that requires full annotations on a subset of images and no annotations on the remaining images \cite{radosavovic2018data,sohn2020detection,liu2021unbiased}, and 
(3) \textit{learning with partial annotations} where only a few objects in images are annotated while the others are not \cite{wu2019soft,zhang2020solving}.
Among them, weakly-supervised learning usually requires a large number of unsupervised object proposals generated with Selective Search~\cite{uijlings2013ss} or EdgeBox~\cite{zitnick2014edgebox}. This leads to longer training and most importantly longer inference time. Training a good weakly supervised detector often needs specifically designed heuristics, and even then the performance is more than 20 points lower than a fully-supervised counterpart~\cite{bilen2016weakly,ren2020ufo2}.
Conventionally, the \textit{partial images} (PI) setting is well studied in image classification~\cite{fixmatch} and object detection~\cite{sohn2020detection,liu2021unbiased}. Recent methods follow the paradigm of online teacher-student distillation model. The teacher model usually generates pseudo labels for the unlabeled image with weak augmentations, and the student model uses the pseudo labels on strongly augmented version of the same image. The teacher model is updated with exponential moving average (EMA) of the student model. Example of this paradigm for object detection includes STAC and Unbiased Teacher~\cite{sohn2020detection,liu2021unbiased}.
These methods have achieved great success in natural image datasets from sparse scenes (average bounding box per image is small) such as VOC~\cite{VOC} and COCO~\cite{COCO}. However, its performance is unproven for small to medium scale and datasets from dense scenes.
In this paper, we focus on learning with partial annotations and show that \AC performs better than (1) SSL with similar annotation budget and (2) existing methods in partial annotations~\cite{wu2019soft,zhang2020solving} that are based on modifications to the loss function and achieve good performance on VOC~\cite{VOC} and OpenImages~\cite{OpenImages}.
For example, Soft Sampling~\cite{wu2019soft} reduces the effect of false negatives by down weighing the gradients of ROIs based on overlap with the ground-truth annotations and Recalibration Loss~\cite{zhang2020solving} modifies the loss of the hard negatives based on prediction scores.

\section{Partial Annotations and ComplETR}
In this section, we introduce the partial annotations setting and the technical principles of the proposed ComplETR.

\subsection{Partial annotations}
Formally, we consider a object detection training set with $n$ samples as $\mathcal{X} = \left\{(X_i,Y_i)\right\}_{i=1}^{n}$, where $X_i$ is the $i^{th}$ training image and $Y_i$ represents the full $m_i$ annotations with $Y_i = \left\{y_j\right\}_{j=1}^{m_i}$. Each $\left\{y_j\right\}$ is a $5$-dim vector containing the bounding box coordinate and instance label. In this paper, we focus on detection in dense scenes, where the average number of annotations per image $A_m = \frac{1}{n}\sum_{i=1}^{n}m_i$ is large. Specifically, we consider cases where $A_m \geq 15$. For comparison, PASCAL VOC~\cite{VOC} has $A_m \approx 2$ and COCO~\cite{COCO} has $A_m \approx 8$.

In order to reduce the annotation cost for a dense dataset, we need to reduce the total annotations $nA_m$. Naturally, this leads to two paths. (1) We reduce the total number of images to annotate from $n$ to $\hat{n}$ with $\hat{n} \ll n$. This leads to the development of semi-supervised and unsupervised learning in the extreme setting. We refer to this setting as \textit{partial images} (PI). (2) We reduce the average annotations per image $A_m$ to $\hat{A}_m$ with $\hat{A}_m \ll A_m$. We refer to this path as \textit{partial annotations} (PA) which is rarely studied in the literature. Most existing methods~\cite{wu2019soft,zhang2020solving} treat this setting as an inherent problem of missing bounding box annotations, rather than a path to reduce the annotation cost.
Note that there is a third path, where we reduce the cost of each annotation. This can be the result of better tooling~\cite{papadopoulos12ec}, using alternative annotations such as points~\cite{chen2021point}, and image-level labels~\cite{bilen2016weakly,gao2018notercnn,ren2020instance,ren2020ufo2}. Our focus is on standard bounding box annotation, and hence we will not discuss this path in detail.

In this paper, we propose to use the \textit{partial annotations} setting to reduce the annotation cost for densely annotated datasets. In previous work~\cite{wu2019soft,zhang2020solving}, partial annotations was seen as a \textit{problem} from the inherent difficulty of annotating bounding boxes exhaustively. For example, soft sampling~\cite{wu2019soft} was proposed to reduce the negative effect of training detectors on partially annotated datasets, such as OpenImages~\cite{OpenImages}. We actually consider partial annotations as a \textit{solution} to reduce the annotation cost.

We further compare both the options -- partial images and partial annotations -- to use a given fixed annotation budget. As we see in the results in Section~\ref{S:experiments}, partial annotations where we use few annotations for all (most) images is superior over partial images where we have all annotations in few images.

\subsection{\AC}
\label{S:completer}
The central idea behind \AC is to leverage existing partial annotations to complete missing annotations, which otherwise would be treated as a part of background.
Formally, a partially annotated dataset is defined as $\mathcal{\hat{X}} = \left\{(X_i,\hat{Y}_i)\right\}_{i=1}^{n}$, where $\hat{Y}_i = \left\{y_j\right\}_{j=1}^{\hat{m}_i}$. Here we keep all (or most) of the training images, but annotate only $\hat{m}_i$ instead of $m_i$ bounding boxes for the $i$-th images, where $\hat{m}_i \ll m_i$.

Intuitively, the completion of missing boxes can be done by learning from partially annotated data $\mathcal{\hat{X}}$, and then predicting on the same dataset to get new annotations. Potentially, this can also fit into the online teacher-student model in the \textit{partial images} setting, where the teacher model predicts pseudo labels from the weakly-augmented partially-annotated images and the student model uses these labels as ground-truth for strongly augmented images.

However, we argue there are two main problems with this naive solution. Firstly, as shown in the literature~\cite{wu2019soft,zhang2020solving}, partial annotations is detrimental to detector learning, especially when the fraction of annotations is small. Thus, the teacher-student model may never be robust enough to detect most objects. Secondly, the current online teacher-student learning paradigm has a key issue. The EMA-updated teacher model means the exact same architecture needs to be used for the teacher and student models. Switching detection architecture from the most studied Faster R-CNN to better/faster/more-suitable one will require significant efforts of adaptation and hyper-parameter search if even possible. This leads to worse generalization ability for datasets other than VOC/COCO and less flexible real-world deployment. Densely annotated datasets can originate from vastly different applications, from detecting apples on trees to ships in drone footage. Different detectors may have advantages in different datasets, or even for different annotation percentage. Therefore, we would like to steer away from the online teacher-model.

To address these two issues, we propose an annotation completer framework that includes (1) Two-stream Deformable-DETR (D-DETR) with query patches as the teacher model to generate pseudo ground-truth labels; see Section~\ref{SS:two-stream}, and (2) Decoupled training, where any student model, regardless of its architecture, can learn from the generated pseudo ground-truths; see Section~\ref{SS:training_workflow}.

\subsection{Two-stream D-DETR with query patches}
\label{SS:two-stream}

\begin{figure}[t!]
\vspace{-10pt}
\small
\centering
\begin{picture}(100,165)
\put(-70,-5){\includegraphics[width=0.7\linewidth]{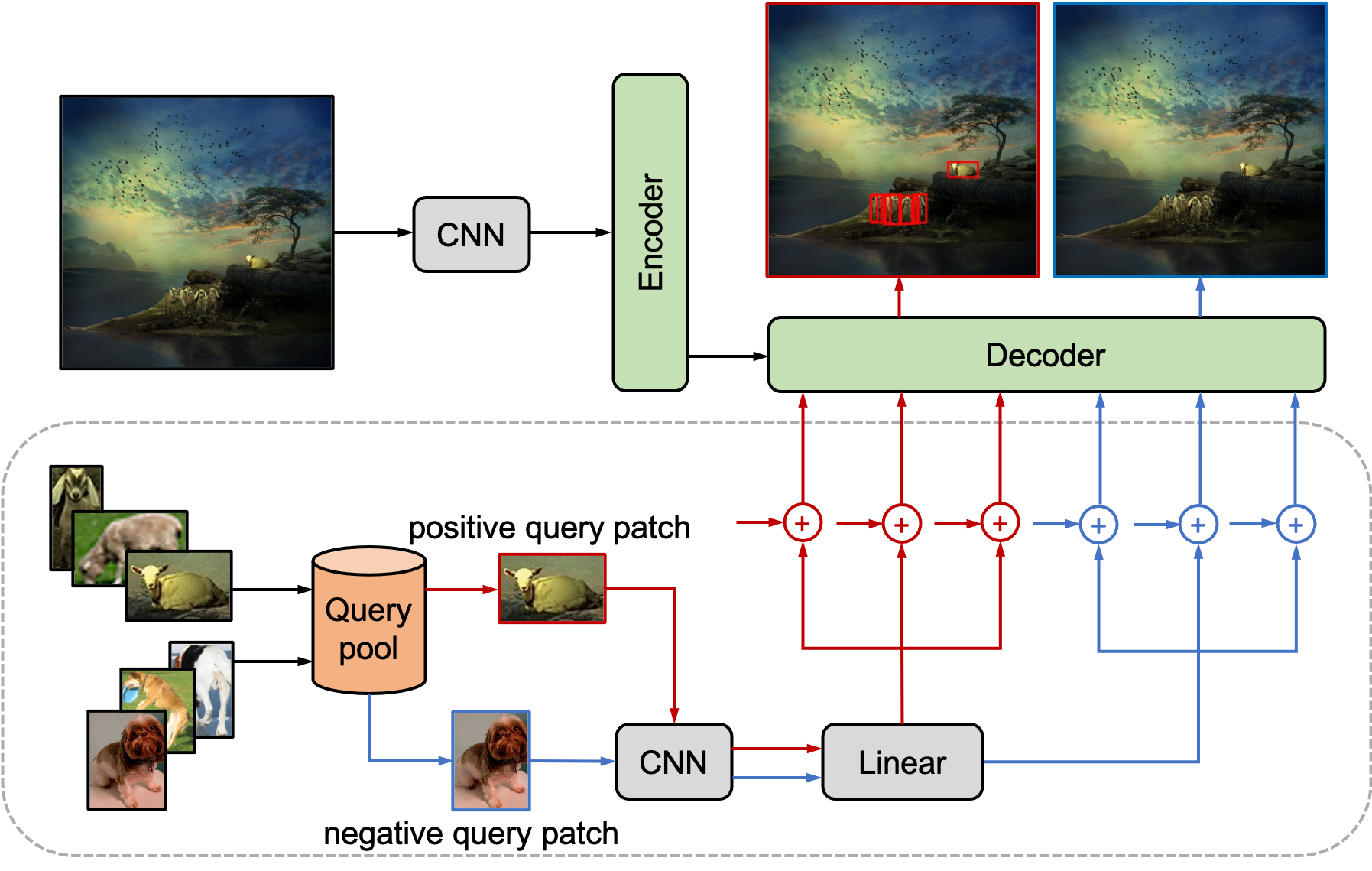}}
\put(60,51){$q_1$}
\put(74,66){$\hat{q}^{+}_1$}
\put(145,51){${q}_N$}
\put(160,66){$\hat{q}^{-}_N$}
\put(75,26){${p}^{+}$}
\put(145,26){${p}^{-}$}
\end{picture}
\caption{\AC uses two streams of inputs: input image and query patches of existing partial annotations. During training, positive and negative query patches are sampled from the query pool, passed through the CNN backbone, and then to the decoder after a linear transformation. The network detects objects corresponding to the class of positive query patches (sheep in this case) while treating negative query patches (dogs in this case) as background.}
\label{F:completer}
\end{figure}
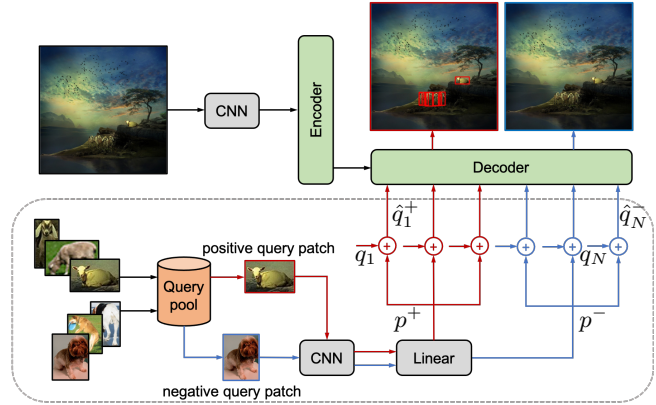

We propose to explicitly use the existing annotations to search for similar matches in the whole image and identify missing objects. To this end, DETR~\cite{carion2020end} and its derivative method such as D-DETR~\cite{zhu2021deformable} have an intriguing property: the cross attention mechanism in the transformer decoder can naturally find correspondences between the image feature maps from the encoder (keys) and the positional encoding of the objects (object queries). Specifically, D-DETR learns $N$ object queries $\{q_i\}_{i=1}^N$ during training, and they are used to attend to different parts of the image and predict $N$ bounding boxes along with their classes. 
We provide additional supervision to D-DETR by augmenting the object queries with the information of the target class, so as to bias the cross-attention modules to make predictions focused on the target class.
Thus, \AC is trained with two-stream inputs: the input image and exemplar query patches from partial annotations.
The overall architecture of Annotation Completer is shown in Figure~\ref{F:completer}; the additional supervision is highlighted in the dashed box.

In particular, query patches are crops of the (partially annotated) ground truth annotations in the input image (or other training images forming a query pool as described in the following paragraph).
The idea of query patches is explored before but in entirely different contexts; e.g.~UP-DETR~\cite{dai2021up} uses random query patch detection as a pretext task for pre-training transformers in an unsupervised way and Meta-DETR~\cite{zhang2021meta} uses patches of the support classes to match them with the queries for meta-learning and few-shot detection.

For pre-processing the query patches, we resize them to a fixed size of $256\times256$ and apply SimCLR-style strong data augmentation~\cite{chen2020simple}.
The pre-processed patches are passed through the shared backbone followed by additional global average pooling (GAP) and linear layers to extract embeddings $p$ given by
\begin{align}
    p = W^{T}\sum_{h,w} \phi(I_p)
    \label{E:embedding}
\end{align}
where, $I_p$ is a query patch with augmentations, $h$, $w$ are the spatial dimensions of the feature map, $\phi$ is the CNN backbone, $C$ is the number of channels in the feature map, and weights $W\in\mathbb{R}^{C\times |q_i|}$ transforms from feature dimension $C$ to the object query dimension $|q_i|, \ i \in \{1,2,\dots,N\}$.
Next, the object queries $q_i$ are augmented with the embedding vectors of the query patches $p$ by a simple addition
\begin{align}
    \hat{q}_i = q_i + p.
    \label{E:augmentation}
\end{align}
The augmented queries $\hat{q}_i, \ i \in \{1,2,\dots,N\}$ are then fed to the decoder and following layers to make predictions.

\textbf{1.~Query pool.}
To identify missing objects with only partial annotations in a target image, query patches from the same image by themselves may not be sufficient.
For example, if the annotated objects in the target image are in the front view, then other objects in the side view may not be recognizable.
Thus, it is helpful to have diverse query patches that can come from other training images.
To this end, we build a \textit{query pool}, which is a collection of cropped patches of all available ground truth annotations in the partially annotated dataset.
Using a query pool clusters the patch features belonging to the same class enabling within-class coherence. We use two types of queries patches, positive and negative. 

\textit{Positive query patches:}
These are the set of query patches belonging to the target class in the query pool which we randomly sample for a positive query. 

\textit{Negative query patches:}
To amplify the focus on the target class, we also sample negative query patches belonging to a non-target class.

For example, consider COCO dataset with 80 classes.
In an image, let's say we have two bounding boxes, one each for Class X and Class Y.
During training, one of the categories is randomly selected as the target class; let's say Class X.
In this case, the patches belonging to Class X are treated as positive query patches and the others belonging to all remaining 79 classes, \textit{including Class Y}, are considered negative query patches.
Following Eq.~\eqref{E:embedding}, we pass positive and negative patches separately through the shared backbone to obtain query embeddings ${p}^{+}$ and ${p}^{-}$, and then apply Eq.~\eqref{E:augmentation} to obtain $\hat{q}^{+}_i$ and $\hat{q}^{-}_i$ queries as the inputs to the transformer decoder.
See an illustration in Figure~\ref{F:completer} with positive and negative query patches of ``sheep" and ``bird", respectively.
The target detections corresponding to the negative queries are set to the background class.
This enables the network to use the patch information effectively to detect objects corresponding to the target class.

\begin{figure}[t!]
\small
\centering

\includegraphics[width=0.3\linewidth]{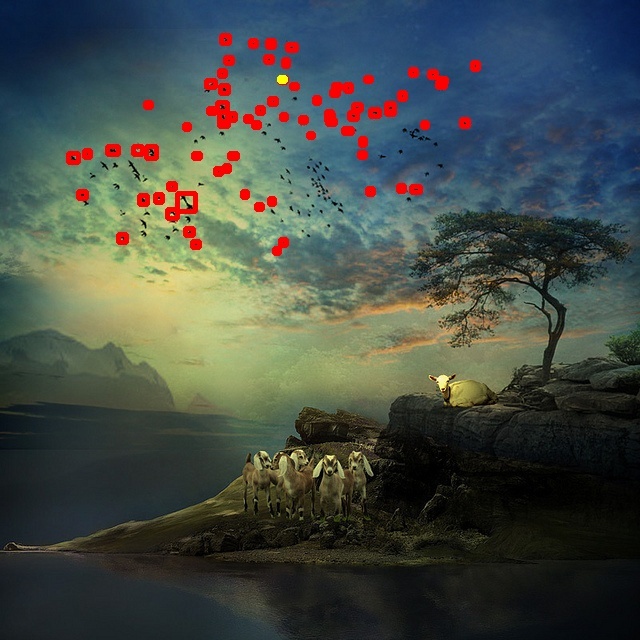}
\includegraphics[width=0.3\linewidth]{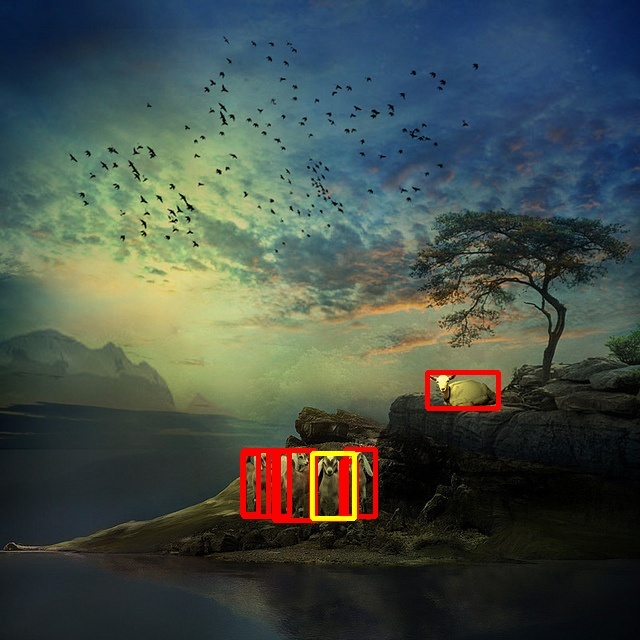}

\caption{Evaluation: Visualizations of detections when different query patches (yellow) from the same image are passed through \AC. Left: query patch of a bird predicts the missing birds. Right: query patch of a sheep predicts the missing sheep.}
\label{F:attention_map}
\end{figure}

\textbf{2.~Binary classification loss.}
\label{SS:binary_loss}
\AC attends to a target class of the available (but partial) annotations in a target image using a binary classification loss that determines the presence or absence of the target class.
Specifically, all ground truth annotations belonging to the target class are labeled as positive ($c=1$), and the others are treated as negative ($c=0$).
Following D-DETR, we choose Focal Loss~\cite{lin2017focal} to compute class probabilities $\hat{p}$ in the classification loss
\begin{align}
\mathcal{L}_{\text{cls}} = - \sum_{i=1}^{N} \log \hat{p}_{\hat{\sigma}(i)}(c_i),
\end{align}
where $N$ denotes the number of detections, $\hat{\sigma}$ the optimal bipartite matching obtained from Hungarian loss, and $\hat{p}_{\hat{\sigma}(i)}(c_i)$ is the probability of the ground truth class $c_i \in \{0,1\}$ for the detection with index $\hat{\sigma}(i)$.
For the detections corresponding to negative queries $\hat{q}^{-}_i$, we set $c_i=0$.
For bounding box losses, we use the standard L1 regression loss and generalized IoU (GIoU) loss~\cite{rezatofighi2019generalized}:
{
\small
\begin{align}
\mathcal{L}_{\text{box}} = \sum_{i|c_i=1} \lambda_{\text{L1}} \lVert b_i - \hat{b}_{\hat{\sigma}(i)} \rVert_1 + \lambda_{\text{GIoU}} \mathcal{L}_{\text{GIoU}}(b_i, \hat{b}_{\hat{\sigma}(i)}),
\end{align}
}%
where $b_i$ and $\hat{b}_i$ denote ground truth and predicted bounding box coordinates, respectively.
Overall, the loss function for training \AC is $\mathcal{L} = \lambda_{\text{cls}} \mathcal{L}_{\text{cls}} + \lambda_{\text{box}} \mathcal{L}_{\text{box}}$.
During evaluation, the binary loss helps \AC generate different detections on the same image based on the query patch as shown in Figure~\ref{F:attention_map}.

\textbf{3.~Soft sampling.}
\label{SS:soft_sampling}
As noted above, \AC uses the standard set-based Hungarian loss in D-DETR for enforcing unique predictions for each ground-truth.
However, when this loss is computed using partial ground-truth annotations, the number of matches inevitably differ from the actual number of objects due to missing annotations.
The detections that have a small overlap (e.g., 0-0.2 IoU) with the ground truth annotations may have been matched to incorrect ground truths because some of the annotations are actually missing.
To reduce the effect of mismatching due to missing annotations, we re-weigh the gradients of the unmatched detections according to their overlap $o_i, \ i \in \{1,2,\dots,N\}$ with the ground truth annotations using soft sampling~\cite{wu2019soft}. The weight $w_i \in [0,1]$ is given by
\begin{align}
    w_i(o_i) = a + (1-a)\text{e}^{-b\text{e}^{-co_i}}.
\end{align}
We adopt the settings for $a$, $b$, and $c$ from \cite{wu2019soft}.
The gradients of the matched detections are not modified, i.e.,~$w_i=1$.
Although soft sampling also down weighs true negatives, we empirically observe that it improves the performance of \AC, especially when the number of missing annotations are large.

\subsection{Decoupled training process}
\label{SS:training_workflow}
To enable the use of any object detector (Step 4 below) for downstream tasks with \AC, we propose a training workflow with five stages, including an optional stage for annotation refinement via multi-stage detector training.

\textbf{1.~Pre-training}: \AC is pre-trained on a large and fully annotated dataset to learn the annotation completion task.
We randomly sample one or multiple ground truth bounding boxes from the same image or other training images as query patches and the full annotations as the ground truth. This way, the model learns to complete annotations during pre-training. Since we choose a binary classification loss (see Section~\ref{SS:binary_loss}), the completion property can generalize to new classes and domains. The pre-training step is done only once since it is common to all target datasets. In our experiments, we use COCO as the pre-training dataset. We initialize the pre-training with self-supervised SwAV~\cite{caron2020swav} representations and freeze the backbone during this step.

\textbf{2.~Partial annotation fine-tuning}: Next, we unfreeze the backbone and fine-tune the pre-trained \AC on a partially annotated target dataset to learn the specific properties of the target dataset. We demonstrate that we can generalize well even to new classes that are absent in the pre-training dataset.

\textbf{3.~Annotation completion}: The fine-tuned \AC predicts missing annotations in the target dataset, which are then merged with the original partial annotations to obtain complete annotations.

\textbf{4.~Detector training}: A separate off-the-shelf object detector like Faster R-CNN, D-DETR, etc., is trained on the target training dataset but with the completed annotations. We show that the performance of a detector can be improved irrespective of its architecture and/or specific loss function.

\textbf{5.~Multi-stage training}: At this step, we can optionally refine the annotations by generating pseudo labels on the target dataset using the final detector and then retraining the detector. We observe that for several datasets, two-stage \AC indeed improves the performance.

\section{Experiments}
\label{S:experiments}

\subsection{Datasets}
\label{SS:datasets}
We use datasets from different domains such as Minneapple~\cite{Minneapple}, Bees~\cite{Bees}, CrowdHuman~\cite{CrowdHuman}, ``LVISdense'' (subsampled from LVIS~\cite{LVIS}), and ``COCOdense'' (subsampled from COCO~\cite{COCO}) as partially annotated target datasets.
For the first three datasets, we use COCO to pre-train all methods for fair comparisons. In the case of ``LVISdense'' and ``COCOdense'', we use ``COCO60'' instead.
``COCO60'' is constructed by removing images from COCO that have any object from the 20 VOC classes.
For Minneapple, we did a 60-40 split of the publicly available annotated (training) dataset containing 670 images to obtain 403 training and 267 test images. For Bees, we constructed a dataset with 1064 and 262 images for training and test, respectively. For CrowdHuman, we used the standard split of 15000 training and 5000 test images. 
For LVISdense, we subsample 13 classes with dense annotations from LVIS that are absent in the COCO60 pre-training set, resulting in 2937 training and 570 test images.
For COCOdense, we subsample 6 classes with dense annotations from COCO that are absent in the COCO60, resulting in 1723 training and 76 test images.
All target datasets have high intra-class similarity and high object density (at least 15 bounding box per image on average).
Note that while ``apple'' in Minneapple and ``person'' in CrowdHuman are present in COCO classes, we also show generalization to new classes via the Bees (absent in COCO), LVISdense (all 13 classes are absent in COCO60), COCOdense (all 6 classes are absent in COCO60).

\subsection{Annotations protocol}
\label{SS:sampling_protocol}
We construct annotations for our experiments as per the following sampling protocol.
For both settings, we obtain X\% annotations with three random seeds and compute the average metrics.
We assume the datasets listed in Section~\ref{SS:datasets} are fully annotated.
We only sample from the training data to generate X\% annotations; the test set is used as is.

\textbf{Partial images.}
To obtain X\% annotations, we keep each image with X\% probability. The selected images are treated as labeled. We use the remaining images as unlabeled data for semi-supervised baselines wherever appropriate.

\textbf{Partial annotations.}
To obtain X\% annotations, we keep each bounding box with X\% probability. 
After the sampling, if an image does not contain any annotations, we drop it from the training set. Hence, 1\%, 5\%, etc., annotation settings can have different number of images.

We evaluate both settings with \textit{similar} annotation budget; it is practically hard to ensure the exact same number of annotations following the above sampling protocol without introducing a sampling bias.


\subsection{Baselines for comparison}
We provide a comprehensive comparison against the baselines in both annotation settings.

\textbf{Partial images.}
In this setting, all objects are annotated but use fewer images for training. This is indeed attractive since it does not suffer from the issue of training on noisy ground-truths.
A simple baseline for comparison is \textit{Pseudo Labeling} where we generate pseudo labels using a detector trained on the annotated images and then re-train the detector; we test pseudo labeling with Faster R-CNN and Deformable DETR.
In the Faster R-CNN framework, we also test the state-of-the-art SSL method for object detection, \textit{Unbiased Teacher}~\cite{liu2021unbiased}, that also uses unlabeled images in the partial images setting. 

\textbf{Partial annotations.}
In this setting, some objects are annotated in most images. Thus, the missing annotations are inevitably treated as background by a detector; i.e.,~the detector trains on some false negatives in the ground-truths.
Existing work that can be applied to the partial annotations setting are based on modification of the loss function. We test \textit{Soft Sampling}~\cite{wu2019soft} which was originally proposed to mitigate the effect of missing annotations in existing datasets by modifying the gradients of ROIs based on overlap with the ground-truth annotations. We also evaluate Background \textit{Re-calibration Loss}~\cite{zhang2020solving} which modifies the loss for confusing anchors (hard negatives) to tackle the problem of missing annotations.
Additionally, we study \textit{Pseudo Labeling} together with \AC in a two-stage fashion as described in Section~\ref{SS:training_workflow}.

\subsection{Experimental settings}
\label{SS:exp_settings}

\textbf{Faster/Cascade-RCNN, CenterNet2, and Unbiased Teacher:}
For fine-tuning detectors on target datasets, we initialize the models with COCO pre-trained weights (or COCO60 for LVIS and COCO dense as noted in Section~\ref{SS:datasets}) and train for at least 1000 iterations and at most 12 epochs.  We use learning rate $5\times10^{-4}$  with SGD and decay by 0.1 at 80\% iterations. We observed that this small learning rate is essential, especially in the case of extreme partial annotations where a large learning rate easily overfits to missing annotations; see an ablation study in Section~\ref{S:learning_rate} (supplementary material).
For all datasets, we show the performance at 100\% annotations with the best learning rate in \{0.0005, 0.001, 0.01, 0.02\}.
For Unbiased Teacher, we fixed the burn-in period as 200 iterations for all scenarios.
In all cases, we tested the official implementations in Detectron2~\cite{wu2019detectron2} with ResNet-50+FPN backbone and default settings unless noted otherwise.
\textbf{D-DETR based methods:} In all cases, we use the default learning rate of $2\times10^{-4}$ for the transformer and $2\times10^{-5}$ for the ResNet-50 backbone, with AdamW optimizer. HPO on other learning rates does not improve the results. We chose the default 300 object queries for both \AC and D-DETR.
For pretraining of \AC, we initialized with SwAV~\cite{caron2020swav} representation and trained for 100 epochs with an LR decay by 0.1 at the 80-th epoch.
For fine-tuning \AC and D-DETR detectors, we ran all experiments for 15 epochs with decay by 0.1 at 10-th epoch.
\textbf{All methods:}
We use a batch size of 16 on 8 GPUs, except in the 1\% partial images setting for Minneapple, Bees, and COCOdense where 1\% images correspond to 4-18 images in which case we use single GPU.
All the detectors and \AC are pretrained on the same dataset before fine-tuning for a fair comparison. \textbf{Threshold selection:} For pseudo labeling and \AC generated annotations we need to identify a threshold to filter out low confident detections. We chose a fixed threshold of 0.7 for the partial images setting. 
This is consistent with the threshold used in Unbiased Teacher in the online student-teacher framework. When \AC is fine-tuned on partial annotations according to Step 2 in Section~\ref{SS:training_workflow}, the confidence scores tend to be lower. Hence, we choose a fixed threshold of 0.3 to filter out low confident completions; refer Step 3 in Section~\ref{SS:training_workflow}. Note that this selection is lower than 0.7 used for pseudo labeling in the partial images setting where scores tend to be higher due to full annotations. See ablation on the choice of thresholds in Section~\ref{S:ablation_thresholds} (supplementary material).

\subsection{Analysis}
\label{SS:analysis}

We use mAP@0.5 IoU (AP50) as the metric for comparison throughout this section.
We present baselines based on Faster R-CNN~\cite{ren2015faster} and 
Deformable DETR~\cite{zhu2021deformable} for both partial images and partial annotations settings in Table~\ref{T:baselines}.
Note that both the settings have similar annotation budget for any given annotation scenario.

\textbf{Partial images.}
Pseudo labeling improves performance in most cases for both Faster R-CNN and D-DETR.
Unbiased Teacher provides marginal improvement in the extreme case of 1\% images for Minneapple and Bees while still under-performing compared to naive pseudo labeling.
Although we tune the learning rate and burn-in period specifically for these datasets, other hyper-parameters/components in Unbiased Teacher may not be optimal for dense object detection. This validates our hypothesis that SSL methods may not perform well on dense datasets without significant modifications.

\textbf{Partial annotations.}
Faster R-CNN suffers more in the presence of missing annotations while D-DETR is relatively robust and performs better in majority of the cases.
Our hypothesis is that sigmoid Focal Loss~\cite{lin2017focal} in D-DETR provides this robustness. When we replaced it with the standard sigmoid binary cross entropy loss, the performance of D-DETR fell steeply; we show an ablation in Section~\ref{S:focal_loss} (supplementary material).
We observe that Soft Sampling~\cite{wu2019soft} improves the performance in most scenarios.
Recalibration Loss~\cite{zhang2020solving} did not improve the performance of Faster R-CNN in our experiments; see results on Minneapple in Figure~\ref{F:examples}.
For the recalibration threshold, a score below which the focal loss is mirrored to reduce the weights on hard negatives, we tried 0.25 besides the default setting of 0.5 optimized for large-scale datasets like VOC/COCO.

\begin{table}[t]
\small
\centering
\caption{Comparison against baselines on Minneapple, Bees, CrowdHuman, LVISdense, and COCOdense. \AC 2S refers to \AC two-stage.}
\label{T:baselines}
\resizebox{1\columnwidth}{!}{
\begin{tabular}{llcccccc}
\toprule
\multirow{2}{*}{Setting}             & \multirow{2}{*}{Method} & \multicolumn{2}{c}{Minneapple}                    & \multicolumn{1}{c}{Bees} & \multicolumn{1}{c}{Crowd} & \multicolumn{1}{c}{LVISdense} & \multicolumn{1}{c}{COCOdense} \\
\cmidrule(lr){3-4} \cmidrule(lr){5-5} \cmidrule(lr){6-6} \cmidrule(lr){7-7} \cmidrule(lr){8-8}
                                     &                         & \multicolumn{1}{c}{1\%} & \multicolumn{1}{c}{5\%} & \multicolumn{1}{c}{1\%}  & \multicolumn{1}{c}{1\%} & \multicolumn{1}{c}{1\%} & \multicolumn{1}{c}{1\%}       \\
\cmidrule(lr){1-1} \cmidrule(lr){2-2} \cmidrule(lr){3-3} \cmidrule(lr){4-4} \cmidrule(lr){5-5} \cmidrule(lr){6-6} \cmidrule(lr){7-7} \cmidrule(lr){8-8}                      
\multirow{5}{*}{\parbox{0.18\linewidth}{partial\\images}}      & Faster R-CNN            & 50.0                    & 64.4                    & 55.3                     & 66.7   &      4.9          &       8.4                        \\
                                     & Pseudo Labeling         & 53.6                    & 65.4                    & 58.9                     & 65.6            &   8.4                     &   18.9         \\
                                     & Unbiased Teacher        & 52.5                    & 62.1                    & 56.1                     & 65.5             &    4.4                    &    8.3        \\
\cmidrule(lr){2-8}                             
                                    & D-DETR                  & 36.6                    & 46.6                    & 58.1                     & 67.6                &   4.2                   &   7.7        \\
                                     & Pseudo Labeling         & 51.5                    & 54.6                    & 70.5                     & 68.4               &      5.6                &    11.7        \\
\midrule                                     
\multirow{7}{*}{\parbox{0.18\linewidth}{partial\\annotations}} & Faster R-CNN            & 43.6                    & 62.1                    & 38.8                     & 59.6                           &   1.3      &      4.6       \\
                                     & Soft Sampling           & 49.1                    & 61.8                    & 44.5                     & 61.9                 &    1.9                &    7.6       \\
                                     & Faster R-CNN+\AC                & \textbf{65.3}           & \textbf{70.3}           & \textbf{70.8}            & \textbf{69.0}  & 7.5            &       17.9           \\
                                     & Faster R-CNN+\AC 2S \ \        & 63.2                    & -                       & \textbf{71.1}            & 67.6               &    6.5         &     \textbf{20.2}       \\
\cmidrule(lr){2-8}
                                    & D-DETR                  & 49.6                    & 56.9                    & 61.6                     & 63.3                     &    7.5             &     17.3      \\
                                     & D-DETR+\AC                & \textbf{64.8}           & \textbf{69.1}           & \textbf{77.9}            & \textbf{70.8}         &      \textbf{10.3}      &      \textbf{23.1}    \\
                                     & D-DETR+\AC 2S        & 63.7                    & -                       & \textbf{79.0}            & \textbf{72.3}               &     \textbf{12.2}    &  \textbf{27.3}  \\
\bottomrule                                     
\end{tabular}
}
\vspace{-15pt}
\end{table}

\textbf{\AC} consistently outperforms all the baselines across both the partial images and partial annotations settings by a wide margin.
We also observe that, in several cases, pseudo labeling is complementary to \AC and using them together in a multi-stage setting (which we refer to as \AC 2-stage or 2S) further improves the performance; see Step 5 in Section~\ref{SS:training_workflow}. We present an ablation study on using pseudo labeling \textit{before} or \textit{after} \AC in Section~\ref{S:ablation_pseudo_labels} (supplementary material).

\begin{table*}[!t]
\small
\centering
\caption{Evaluation of performances for different detectors on partial and completed annotations on Minneapple, Bees, CrowdHuman, LVISdense, and COCOdense.}
\label{T:detectors}
\resizebox{\columnwidth}{!}{
\begin{tabular}{lccccccc}
\toprule
\multirow{2}{*}{Dataset}    & \multicolumn{2}{c}{\multirow{2}{*}{\parbox{0.18\linewidth}{Amount of\\annotations}}} & \multicolumn{2}{c}{Faster R-CNN}                                  & \multicolumn{1}{c}{Cascade R-CNN} & \multicolumn{1}{c}{CenterNet2} & \multicolumn{1}{c}{D-DETR} \\
\cmidrule(lr){4-5} \cmidrule(lr){6-6} \cmidrule(lr){7-7} \cmidrule(lr){8-8}
                            & \multicolumn{2}{c}{}                                      & \multicolumn{1}{c}{R50+FPN} & \multicolumn{1}{c}{+ soft sampling} & \multicolumn{1}{c}{R50+FPN}       & \multicolumn{1}{c}{R50+FPN}    & \multicolumn{1}{c}{R50}    \\
\cmidrule(lr){1-1} \cmidrule(lr){2-3} \cmidrule(lr){4-4} \cmidrule(lr){5-5} \cmidrule(lr){6-6} \cmidrule(lr){7-7} \cmidrule(lr){8-8}                           
\multirow{5}{*}{Minneapple} & \multirow{2}{*}{1\%}              & partial               & 43.6                        & 49.1                                & 49.9                              & 40.1                           & 49.6                       \\
                            &                                   & completed             & \textbf{65.3}               & \textbf{65.7}                       & \textbf{65.4}                     & \textbf{66.0}                  & \textbf{64.8}              \\
\cmidrule(lr){2-2} \cmidrule(lr){3-3}                            
                            & \multirow{2}{*}{5\%}              & partial               & 62.1                        & 61.8                                & 59.7                              & 61.8                           & 56.9                       \\
                            &                                   & completed             & \textbf{70.3}                        & \textbf{68.7 }                               & \textbf{69.0}                              & \textbf{69.2}                           & \textbf{69.1}                       \\
\cmidrule(lr){2-2} \cmidrule(lr){3-3}                            
                            & 100\%                             & full                  & 70.5                        & 68.6                                & 67.5                              & 67.7                           & 67.7                       \\
\midrule
\multirow{3}{*}{Bees}       & \multirow{2}{*}{1\%}              & partial               & 38.8                        & 44.5                                & 44.5                              & 39.0                           & 61.6                       \\
                            &                                   & completed             & \textbf{70.8}               & \textbf{74.0}                                & \textbf{71.1}                              & \textbf{72.1}                           & \textbf{77.9}
                            \\
\cmidrule(lr){2-2} \cmidrule(lr){3-3}                            
                            & 100\%                             & full                  & 85.6                        & 85.2                                & 87.6                              & 88.7                           & 93.6                       \\
\midrule
\multirow{3}{*}{Crowd} & \multirow{2}{*}{1\%}              & partial               & 59.6                        & 61.9                                & 63.9                              & 63.5                           & 63.3                       \\
                            &                                   & completed             & \textbf{69.0}               & \textbf{68.0}                           & \textbf{69.8}                     & \textbf{68.8}                  & \textbf{70.8}              \\
\cmidrule(lr){2-2} \cmidrule(lr){3-3}                            
                            & 100\%                             & full                  & 72.3                        & 71.7                                & 73.0                              & 71.5                           & 82.7 \\
\midrule
\multirow{3}{*}{LVISdense} & \multirow{2}{*}{1\%}              & partial               & 1.3                        & 1.9                                & 2.0                              & 2.7                           & 7.5                       \\
                            &                                   & completed             & \textbf{7.5}               & \textbf{7.8}                           & \textbf{7.2}                     & \textbf{7.1}                  & \textbf{10.3}              \\
\cmidrule(lr){2-2} \cmidrule(lr){3-3}                            
                            & 100\%                             & full                  & 23.9                        &    23.5                             &          24.6                     &    24.3                        & 28.0 \\
\midrule
\multirow{3}{*}{COCOdense} & \multirow{2}{*}{1\%}              & partial               & 4.6                        & 7.6                                & 4.6                              & 6.3                           & 17.3                       \\
                            &                                   & completed             & \textbf{17.9}               & \textbf{19.7}                           & \textbf{16.5}                     & \textbf{18.0}                  & \textbf{23.1}              \\
\cmidrule(lr){2-2} \cmidrule(lr){3-3}                            
                            & 100\%                             & full                  & 47.4                        & 45.6                                & 46.9                              & 47.0                           & 48.5 \\
\bottomrule
\end{tabular}
}
\end{table*}

Since different detectors may have advantages in different applications, we show an elaborate comparison on the flexibility of choosing down-stream detector with the decoupled offline training (Step 4 in Section~\ref{SS:training_workflow}) in Table~\ref{T:detectors}.
Here ``partial'' refers to annotations obtained by the sampling protocol in Section~\ref{SS:sampling_protocol}, ``completed'' refers to annotations obtained by \AC, and ``full'' to the original 100\% annotations. By generating high quality ground-truths of missing annotations, \AC improves the performance of any detector such as Faster R-CNN, Cascade R-CNN, CenterNet2, and Deformable DETR, irrespective of its design and architecture. In the case of Minneapple, Faster R-CNN on completed annotations achieves AP50@100\% annotations with only ~5\% annotations.
\AC is also complementary to a strong baseline like Soft Sampling and improves its performance further. In the column for Soft Sampling, we apply it in the loss function of both \AC (Section~\ref{SS:soft_sampling}) and Faster R-CNN.

Overall, our results show that, through annotation completion, annotating datasets partially is actually a promising path to reducing the cost of annotation for detection in dense scenes.

\section{Ablation studies}
We present six ablation studies, all on Minneapple dataset. Refer to the supplementary material for details on each of them.

\noindent\textbf{1.~Threshold selection.}
After \AC is fine-tuned on partial annotations according to Step 2 in Section~\ref{SS:training_workflow}, we need a threshold to filter noisy completions in Step 3 of Section~\ref{SS:training_workflow}. We fix a threshold of 0.3 based on the experiments; see details in Section~\ref{S:ablation_thresholds}.

\noindent\textbf{2.~SwAV pre-training.}
In Table~\ref{T:baselines}, \textit{\AC} used with Faster R-CNN or D-DETR uses SwAV pre-training; see Step1 in Section~\ref{SS:training_workflow}. We observe that self-supervised training is beneficial to \AC; we analyze its effect on \AC and D-DETR detector in Section~\ref{S:ablation_completer}.

\noindent\textbf{3.~Positive and negative query patches.}
A query pool with both positive and negative query patches plays an important role in learning the annotation completion task. See details in Section~\ref{SS:two-stream} and an ablation study in Section~\ref{S:negative_patches}.

\noindent\textbf{4.~Multi-stage \AC}.
When pseudo labeling is used \textit{after} annotation completion (\AC 2S), the performance further improves in several cases; see Step 5 in Section~\ref{SS:training_workflow} and empirical results in Table~\ref{T:baselines}. Alternatively, we can use pseudo labeling \textit{before} completion. We provide an ablation study in Section~\ref{S:ablation_pseudo_labels} to show that the former setting is better.

\noindent\textbf{5.~Focal loss for partial annotations.}
We empirically show that focal loss in D-DETR is critical to its better performance over Faster R-CNN in the partial annotations setting. See an ablation study Section~\ref{S:focal_loss}.

\noindent\textbf{6.~Learning rate.}
Finally, we provide an ablation on the choice of learning rate for tuning the baselines before fine-tuning on partial annotations in Section~\ref{S:learning_rate}.

\section{Conclusion}
We propose partial annotations as a means to reduce annotation cost for dense object detection datasets. We introduce \AC that uses intra-class similarity across images to find missing annotations previously un-annotated by the user. We show that through annotation completion, \AC outperforms all existing methods in partial annotations as well as semi-supervised method that requires similar annotation budget but need full annotations in fewer images. \AC generates new ground-truth annotations that can be used to improve performance of any off-the-shelf object detector regardless of its architecture.

\clearpage
%
%
\bibliographystyle{splncs04}
\bibliography{ms}

\begin{thebibliography}{10}
\providecommand{\url}[1]{\texttt{#1}}
\providecommand{\urlprefix}{URL }
\providecommand{\doi}[1]{https://doi.org/#1}

\bibitem{Bees}
BC, L.: Noninvasive bee tracking in videos: deep learning algorithms and cloud
  platform design specifications.
  \url{https://lila.science/datasets/boxes-on-bees-and-pollen} (2021)

\bibitem{bilen2016weakly}
Bilen, H., Vedaldi, A.: Weakly supervised deep detection networks. In:
  Proceedings of the IEEE Conference on Computer Vision and Pattern Recognition
  (CVPR) (2016)

\bibitem{cai2018cascadercnn}
Cai, Z., Vasconcelos, N.: {Cascade R-CNN}: Delving into high quality object
  detection. In: Proceedings of the IEEE Conference on Computer Vision and
  Pattern Recognition (CVPR) (2018)

\bibitem{carion2020end}
Carion, N., Massa, F., Synnaeve, G., Usunier, N., Kirillov, A., Zagoruyko, S.:
  End-to-end object detection with transformers. In: Proceedings of the
  European Conference on Computer Vision (ECCV) (2020)

\bibitem{caron2020swav}
Caron, M., Goyal, P., Misra, I., Mairal, J., Bojanowski, P., Joulin, A.:
  Unsupervised learning of visual features by contrasting cluster assignments.
  In: Proceedings of the Thirty-fourth Conference on Neural Information
  Processing Systems (NeurIPS) (2020)

\bibitem{chen2021point}
Chen, L., Yang, T., Zhang, X., Zhang, W., Sun, J.: Points as queries: Weakly
  semi-supervised object detection by points. In: {IEEE} CVPR. pp. 8823--8832
  (2021)

\bibitem{chen2020simple}
Chen, T., Kornblith, S., Norouzi, M., Hinton, G.: A simple framework for
  contrastive learning of visual representations. In: Proceedings of the 37th
  International Conference on Machine Learning (ICML) (2020)

\bibitem{dai2021up}
Dai, Z., Cai, B., Lin, Y., Chen, J.: {UP-DETR}: Unsupervised pre-training for
  object detection with transformers. In: Proceedings of the IEEE Conference on
  Computer Vision and Pattern Recognition (CVPR) (2021)

\bibitem{duan2019centernet}
Duan, K., Bai, S., Xie, L., Qi, H., Huang, Q., Tian, Q.: Centernet: Keypoint
  triplets for object detection. In: Proceedings of the International
  Conference on Computer Vision (ICCV) (2019)

\bibitem{VOC}
Everingham, M., Van~Gool, L., Williams, C.K., Winn, J., Zisserman, A.: {The
  pascal visual object classes (VOC) challenge}. IJCV  \textbf{88}(2),
  303--338 (2010)

\bibitem{gao2018notercnn}
Gao, J., Wang, J., Dai, S., Li, L.J., Nevatia, R.: {NOTE-RCNN}: Noise tolerant
  ensemble rcnn for semi-supervised object detection. arXiv:1812.00124  (2018)

\bibitem{SKU110K}
Goldman, E., Herzig, R., Eisenschtat, A., Goldberger, J., Hassner, T.: Precise
  detection in densely packed scenes. In: Proceedings of the IEEE Conference on
  Computer Vision and Pattern Recognition (CVPR) (2019)

\bibitem{LVIS}
Gupta, A., Dollar, P., Girshick, R.: {LVIS}: A dataset for large vocabulary
  instance segmentation. In: Proceedings of the IEEE Conference on Computer
  Vision and Pattern Recognition (CVPR) (2019)

\bibitem{Minneapple}
H{\"a}ni, N., Roy, P., Isler, V.: Minneapple: A benchmark dataset for apple
  detection and segmentation. arXiv:1909.06441  (2019)

\bibitem{CARPK}
Hsieh, M.R., Lin, Y.L., Hsu, W.H.: Drone-based object counting by spatially
  regularized regional proposal networks. In: Proceedings of the IEEE
  International Conference on Computer Vision (ICCV) (2017)

\bibitem{OpenImages}
Kuznetsova, A., Rom, H., Alldrin, N., Uijlings, J., Krasin, I., Pont-Tuset, J.,
  Kamali, S., Popov, S., Malloci, M., Duerig, T., Ferrari, V.: {The Open Images
  Dataset V4: Unified image classification, object detection, and visual
  relationship detection at scale}. arXiv:1811.00982  (2018)

\bibitem{lin2017focal}
Lin, T.Y., Goyal, P., Girshick, R., He, K., Doll{\'a}r, P.: Focal loss for
  dense object detection. In: Proceedings of the IEEE International Conference
  on Computer Vision (ICCV). pp. 2980--2988 (2017)

\bibitem{COCO}
Lin, T.Y., Maire, M., Belongie, S., Bourdev, L., Girshick, R., Hays, J.,
  Perona, P., Ramanan, D., Zitnick, C.L., Dollár, P.: {Microsoft COCO: Common
  Objects in Context}. arXiv:1405.0312  (2014)

\bibitem{liu2021unbiased}
Liu, Y.C., Ma, C.Y., He, Z., Kuo, C.W., Chen, K., Zhang, P., Wu, B., Kira, Z.,
  Vajda, P.: Unbiased teacher for semi-supervised object detection. In:
  Proceedings of the International Conference on Learning Representations
  (ICLR) (2021)

\bibitem{papadopoulos12ec}
Papadopoulos, D.P., Uijlings, J.R.R., Keller, F., Ferrari, V.: Extreme clicking
  for efficient object annotation. In: {IEEE} CVPR. pp. 4940--4949 (2017)

\bibitem{radosavovic2018data}
Radosavovic, I., Dollár, P., Girshick, R., Gkioxari, G., He, K.: Data
  distillation: Towards omni-supervised learning. In: Proceedings of the IEEE
  Conference on Computer Vision and Pattern Recognition (CVPR) (2018)

\bibitem{ren2015faster}
Ren, S., He, K., Girshick, R.B., Sun, J.: Faster r-cnn: Towards real-time
  object detection with region proposal networks. In: Proceedings of the
  Twenty-ninth Conference on Neural Information Processing Systems (NeurIPS)
  (2015)

\bibitem{ren2020instance}
Ren, Z., Yu, Z., Yang, X., Liu, M.Y., Lee, Y.J., Schwing, A.G., Kautz, J.:
  Instance-aware, context-focused, and memory-efficient weakly supervised
  object detection. In: Proceedings of the IEEE Conference on Computer Vision
  and Pattern Recognition (CVPR) (2020)

\bibitem{ren2020ufo2}
Ren, Z., Yu, Z., Yang, X., Liu, M.Y., Schwing, A.G., Kautz, J.: {UFO}$^2$: A
  unified framework towards omni-supervised object detection. In: Proceedings
  of the European Conference on Computer Vision (ECCV) (2020)

\bibitem{rezatofighi2019generalized}
Rezatofighi, H., Tsoi, N., Gwak, J., Sadeghian, A., Reid, I., Savarese, S.:
  Generalized intersection over union: A metric and a loss for bounding box
  regression. In: Proceedings of the IEEE Conference on Computer Vision and
  Pattern Recognition (CVPR). pp. 658--666 (2019)

\bibitem{CrowdHuman}
Shao, S., Zhao, Z., Li, B., Xiao, T., Yu, G., Zhang, X., Sun, J.: Crowdhuman: A
  benchmark for detecting human in a crowd. arXiv preprint arXiv:1805.00123
  (2018)

\bibitem{fixmatch}
Sohn, K., Berthelot, D., Carlini, N., Zhang, Z., Zhang, H., Raffel, C., Cubuk,
  E.D., Kurakin, A., Li, C.: Fixmatch: Simplifying semi-supervised learning
  with consistency and confidence. In: Larochelle, H., Ranzato, M., Hadsell,
  R., Balcan, M., Lin, H. (eds.) NeuRIPS (2020)

\bibitem{sohn2020detection}
Sohn, K., Zhang, Z., Li, C.L., Zhang, H., Lee, C.Y., Pfister, T.: A simple
  semi-supervised learning framework for object detection. arXiv:2005.04757
  (2020)

\bibitem{hao2012crowdsourcing}
Su, H., Deng, J., Li, F.F.: Crowdsourcing annotations for visual object
  detection. In: Proceedings of the AAAI Technical Report, 4th Human
  Computation Workshop (2012)

\bibitem{uijlings2013ss}
Uijlings, J.R.R., van~de Sande, K.E.A., Gevers, T., Smeulders, A.W.M.:
  Selective search for object recognition. International Journal of Computer
  Vision  \textbf{104}(2),  154--171 (2013)

\bibitem{wu2019detectron2}
Wu, Y., Kirillov, A., Massa, F., Lo, W.Y., Girshick, R.: Detectron2.
  \url{https://github.com/facebookresearch/detectron2} (2019)

\bibitem{wu2019soft}
Wu, Z., Bodla, N., Singh, B., Najibi, M., Chellappa, R., Davis, L.S.: Soft
  sampling for robust object detection. In: Proceedings of the British Machine
  Vision Conference (BMVC) (2019)

\bibitem{zhang2021meta}
Zhang, G., Luo, Z., Cui, K., Lu, S.: {Meta-DETR}: Image-level few-shot object
  detection with inter-class correlation exploitation. arXiv:2103.11731  (2021)

\bibitem{zhang2020solving}
Zhang, H., Chen, F., Shen, Z., Hao, Q., Zhu, C., Savvides, M.: Solving
  missing-annotation object detection with background recalibration loss.
  arXiv:2002.05274  (2020)

\bibitem{zhu2021deformable}
Zhu, X., Su, W., Lu, L., Li, B., Wang, X., Dai, J.: {Deformable DETR}:
  Deformable transformers for end-to-end object detection. In: Proceedings of
  the Ninth International Conference on Learning Representations (ICLR) (2021)

\bibitem{zitnick2014edgebox}
Zitnick, C.L., Doll{\'{a}}r, P.: Edge boxes: Locating object proposals from
  edges. In: Proceedings of the European Conference on Computer Vision (ECCV)
  (2014)

\end{thebibliography}

\clearpage
\appendix
\section*{Supplementary material}
We show all ablation studies on Minneapple dataset.

\section{Threshold selection}
\label{S:ablation_thresholds}
After \AC is fine-tuned on partial annotations according to Step 2 in Section~\ref{SS:training_workflow}, we need a threshold to filter noisy completions in Step 3 of Section~\ref{SS:training_workflow}. We fix a threshold of 0.3 based on the experiments in Table~\ref{T:ablation_thresholds} where we show the performance of downstream detectors for several thresholds $\in$ $\{$0.1, 0.2, 0.3, 0.4, 0.5, 0.6$\}$. A threshold of 0.3 shows the best performance for all detectors, except for D-DETR in which case the best performance is not far from the performance at the chosen threshold.
Note that the threshold selected for \AC is lower than the threshold of 0.7 chosen for the partial images setting (following Unbiased Teacher~\cite{liu2021unbiased}) because the confidence scores tend to be lower due to fine-tuning on partial annotations. Figure~\ref{F:scores} shows the score distribution of Faster R-CNN trained on full annotations (partial images setting) vs \AC trained on partial annotations. The cumulative distribution CDF@0.3 for partial annotations is approximately equal to CDF@0.7 for full annotations. Thus, \AC with a threshold of 0.3 selects approximately the same fraction of bounding boxes as obtained with a threshold of 0.7 in the partial images setting.
\begin{minipage}[!b]{\textwidth}
  \begin{minipage}{0.6\textwidth}
    \small
    \centering
    \label{T:ablation_thresholds}
    \resizebox{0.85\textwidth}{!}{
    \begin{tabular}{ccccc}
        \toprule
        \multirow{2}{*}{Threshold} & Faster        & Cascade       & \multirow{2}{*}{CenterNet2} & \multirow{2}{*}{D-DETR} \\
                                   & R-CNN         & R-CNN         &                             &                         \\
        \cmidrule(lr){1-1} \cmidrule(lr){2-2} \cmidrule(lr){3-3} \cmidrule(lr){4-4} \cmidrule(lr){5-5}                      
        0.1                        & 64.8          & 64.3          & 64.6                        & 64.5                    \\
        0.2                        & 65.1          & 65.1          & 65.5                        & 64.7                    \\
        0.3                        & \textbf{65.3} & \textbf{65.4} & \textbf{66.0}               & 64.8                    \\
        0.4                        & 64.2          & 64.3          & 65.7                        & \textbf{65.0}           \\
        0.5                        & 62.2          & 62.2          & 64.4                        & 63.3                    \\
        0.6                        & 60.4          & 59.3          & 62.3                        & 60.5                   \\
        \bottomrule
        \end{tabular}
        }
      \captionof{table}{Performance of different detectors as the threshold to select the completed annotations for \AC is varied.}
    \end{minipage}
    \begin{minipage}{0.3\textwidth}
        \centering
        \includegraphics[width=1\textwidth]{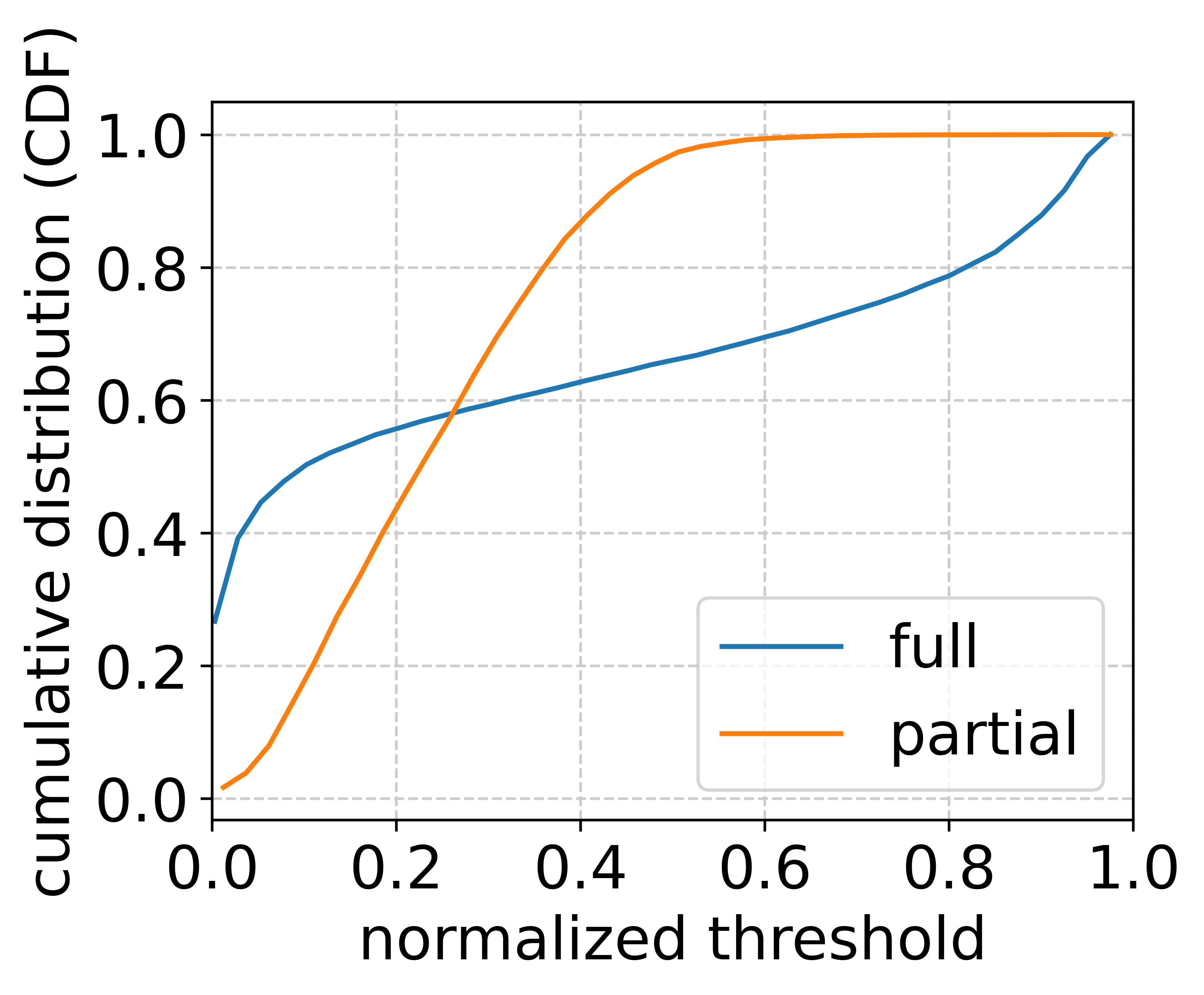}
        \captionof{figure}{Score distribution for training on partial vs full annotations.}
        \label{F:scores}
    \end{minipage}    
  \end{minipage}

\section{Effect of self-supervised pretraining}
\label{S:ablation_completer}

\begin{table}[!t]
\small
\centering
\caption{Ablation study on the quality of completed annotations on the training images (with partial annotations).}
\label{T:ablation_completion}
\begin{tabular}{lcc}
\toprule
Quality of completion & 1\%   & 5\%   \\
\midrule
Partial annotations   & 0.7   & 2.7   \\
\AC w/o fine-tuning   & 43.0  & 43.0  \\
\AC fine-tuned     & 47.3  & 69.3  \\
\AC w/ SwAV    & 58.5  & 73.9  \\
Full annotations      & 100.0 & 100.0  \\
\bottomrule
\end{tabular}
\end{table}

\begin{table}[!t]
\small
\centering
\caption{Ablation study on the effect of SwAV pretraining in \AC on the performance of Faster R-CNN and D-DETR.}
\label{T:ablation_detectors}
\begin{tabular}{lccc}
\toprule
Method & \multicolumn{1}{c}{1\%} & \multicolumn{1}{c}{5\%} & \multicolumn{1}{r}{100\%} \\
\midrule
Faster R-CNN         & 50.0                    & 64.4                    & \multicolumn{1}{r}{70.5}  \\
Faster R-CNN + \AC    & 58.9                    & 68.5                    & -                         \\
Faster R-CNN + \AC w/ SwAV   & \textbf{65.3}           & \textbf{70.3}           & -                         \\
\midrule
D-DETR               & 50.9                    & 56.6                    & \multicolumn{1}{r}{62.5}  \\
D-DETR + SwAV   & 49.6                    & 56.9                    & \multicolumn{1}{r}{67.7}  \\
D-DETR + \AC  & 58.4                    & 64.9                    & -                         \\
D-DETR + \AC w/ SwAV   & \textbf{64.8}           & \textbf{69.1}           & -   \\
\bottomrule
\end{tabular}
\end{table}

In Table~\ref{T:ablation_completion}, we evaluate the quality of completed bounding box annotations obtained from different variations of \AC.
Row 1 corresponding to partial annotations shows AP50 obtained by comparing partial annotations to full annotations.
If we skip the Step 2 of fine-tuning \AC on partial annotations (refer  Section~\ref{SS:training_workflow}), and directly follow Step 3 to complete the missing annotations, we still get a strong completion performance with an AP50 of 43.0. We refer to this as \textit{\AC without fine-tuning}.
As for pretraining in Step 1, we evaluate two choices for initialization -- supervised and self-supervised (SwAV in this case).
In both the cases, we follow Step 2 to fine-tune on the partially annotated dataset and then Step 3 to complete the missing annotations.
These numbers are shown as \textit{\AC fine-tuned} (or simply \AC) and \textit{\AC with SwAV}, respectively.
While supervised initialization improves the completion after fine-tuning, SwAV initialization provides the best completion.
Table~\ref{T:ablation_detectors} shows the effect of SwAV initialization in \AC on the detector trained on the completed annotations.
We again observe the same trend; SwAV initialization improves the downstream performance of both Faster-RCNN and D-DETR.
For reference, we also show the effect of SwAV initialization when we train D-DETR directly on partial annotations (\textit{D-DETR with SwAV}).

\section{Effect of positive and negative query patches}
\label{S:negative_patches}

We show the effect of query pool with positive and negative query patches on the quality of the completed annotations in Table~\ref{T:ablation_negative_patches}.
Row 1 with ``no pool" refers to using ground truth annotations from the target image itself as a query patch; thus we do not need a query pool. Row 2 uses positive query patches from the query pool. Row 2 uses both positive and negative query patches from the query pool. Row 3 combines query pool with SwAV pretraining.

\begin{table}[!t]
\small
\centering
\caption{Effect of positive and negative query patches in \AC.}
\label{T:ablation_negative_patches}
\begin{tabular}{lc}
\toprule
Quality of completion & 1\% \\
\midrule
no query pool   & 47.3 \\
+ positive query patches \ \   & 52.7 \\
+ negative query patches \ \   & 54.5 \\
+ SwAV pretraining    & 58.5 \\
\bottomrule
\end{tabular}
\vspace{-15pt}
\end{table}

\section{Combining pseudo labels with \AC}
\label{S:ablation_pseudo_labels}

\begin{table}[!b]
\small
\centering
\caption{Effect of combining Pseudo Labeling with \AC (before and after).}
\label{T:ablation_pseudo_labels}
\begin{tabular}{lrr}
\toprule
Method                & \multicolumn{1}{c}{D-DETR} & \multicolumn{1}{c}{+ SwAV pretraining} \\
\midrule
1\% annotations & 50.9                        & 49.6                                   \\
\midrule
+ (1) Pseudo Labeling         & 57.1                        & 57.8                                   \\
+ (2) \AC               & 54.0                        & 59.2                                   \\
\midrule
+ (1) \AC               & 58.4                        & 64.8                                   \\
+ (2) Pseudo Labeling         & 59.0                        & 63.7                                   \\
\midrule
100\% annotations & 62.5                        & 67.7                    \\
\bottomrule
\end{tabular}
\end{table}

In Section~\ref{SS:training_workflow}, we proposed an optional step of combining the \AC with Pseudo Labeling. This can be done in two ways:
(a) \textbf{Partial annotations $\rightarrow$ (1) Pseudo Labeling $\rightarrow$ (2) \AC}: train a detector on partial annotations in Stage (1), generate pseudo labels using the detector, train \AC on pseudo labels in Stage (2), obtain completed annotations, and finally train the detector again on the completed annotations, and
(b) \textbf{Partial annotations $\rightarrow$ (1) \AC $\rightarrow$ (2) Pseudo Labeling}: train \AC on partial annotations in Stage (1), obtain completed annotations, train a detector on completed annotations, generate pseudo labels using the detector in Stage (2), and finally train the detector again on the pseudo labels.
For the ablation study, we choose a fixed threshold of 0.3 for the first stage and 0.7 for the second stage.
The scores in the first stage are lower due to partial annotations; we observe 0.3 is a good choice as evident in Table~\ref{T:ablation_thresholds}.
In the second stage, we alleviate the problem of missing annotations to a large extent. Hence, a higher threshold is more reasonable; this is in line with the choice of threshold for full annotations.
We show the results for 1\% annotations  on Minneapple in Table~\ref{T:ablation_pseudo_labels} where \AC after Stage (1) shows better performance as compared to Pseudo Labeling after Stage (1).
We observe Pseudo Labeling on top of completer in Stage (2) can be helpful; we see this pattern for other datasets in Table~\ref{T:baselines} of the main paper.
A large drop in performance in fine-tuning \AC on noisy pseudo labels is due to many false positives. It may help to increase the threshold of Pseudo Labeling in the first stage to keep only high quality completions.

\section{Effect of focal loss on partial annotations}
\label{S:focal_loss}

\begin{table}[!t]
\small
\centering
\caption{Effect of focal loss in D-DETR on partial annotations.}
\label{T:ablation_focal}
\begin{tabular}{lcc}
\toprule
                     & 1\%  & 5\%  \\
\midrule
Sigmoid BCE & 22.5 & 50.7 \\
+ Focal loss~\cite{lin2017focal} & \textbf{50.9} & \textbf{56.6} \\
\bottomrule
\end{tabular}
\end{table}

We observe that focal loss in D-DETR is critical to its better performance on partial annotations (compared to Faster R-CNN). Table~\ref{T:ablation_focal} shows that the performance of D-DETR drops significantly when Focal loss is replaced with the standard Sigmoid Binary Cross Entropy (BCE) loss, i.e., we set $\gamma=0$ in Focal loss. The difference between Focal loss and Sigmoid BCE is lower when the partial annotations are less extreme, for example, compare the case of 1\% annotations with 5\% annotations in Table~\ref{T:ablation_focal}.

\section{Learning rate for partial annotations}
\label{S:learning_rate}

\begin{figure}[!h]
\small
\centering
\includegraphics[width=0.49\linewidth]{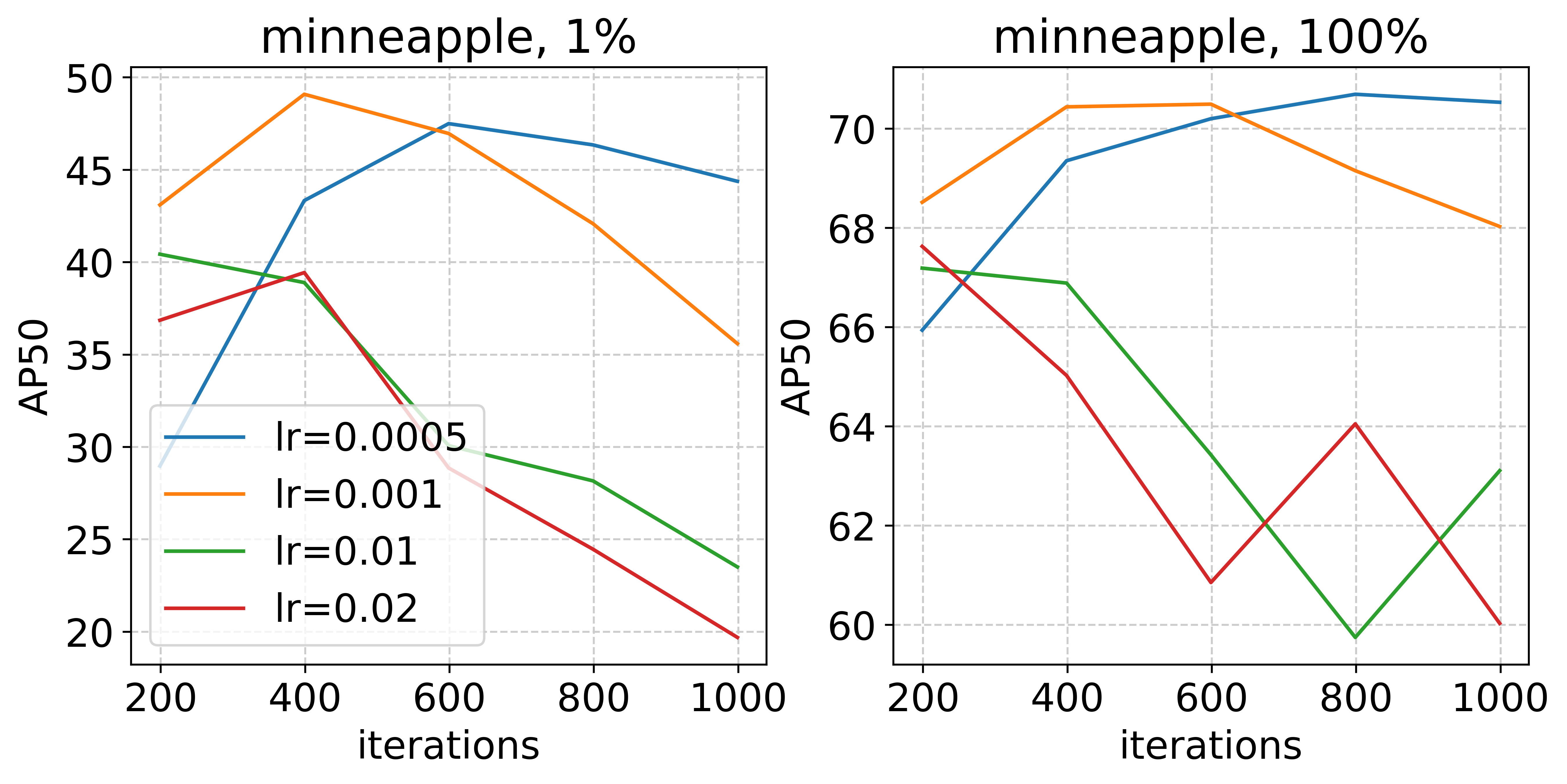}
\hfill
\includegraphics[width=0.49\linewidth]{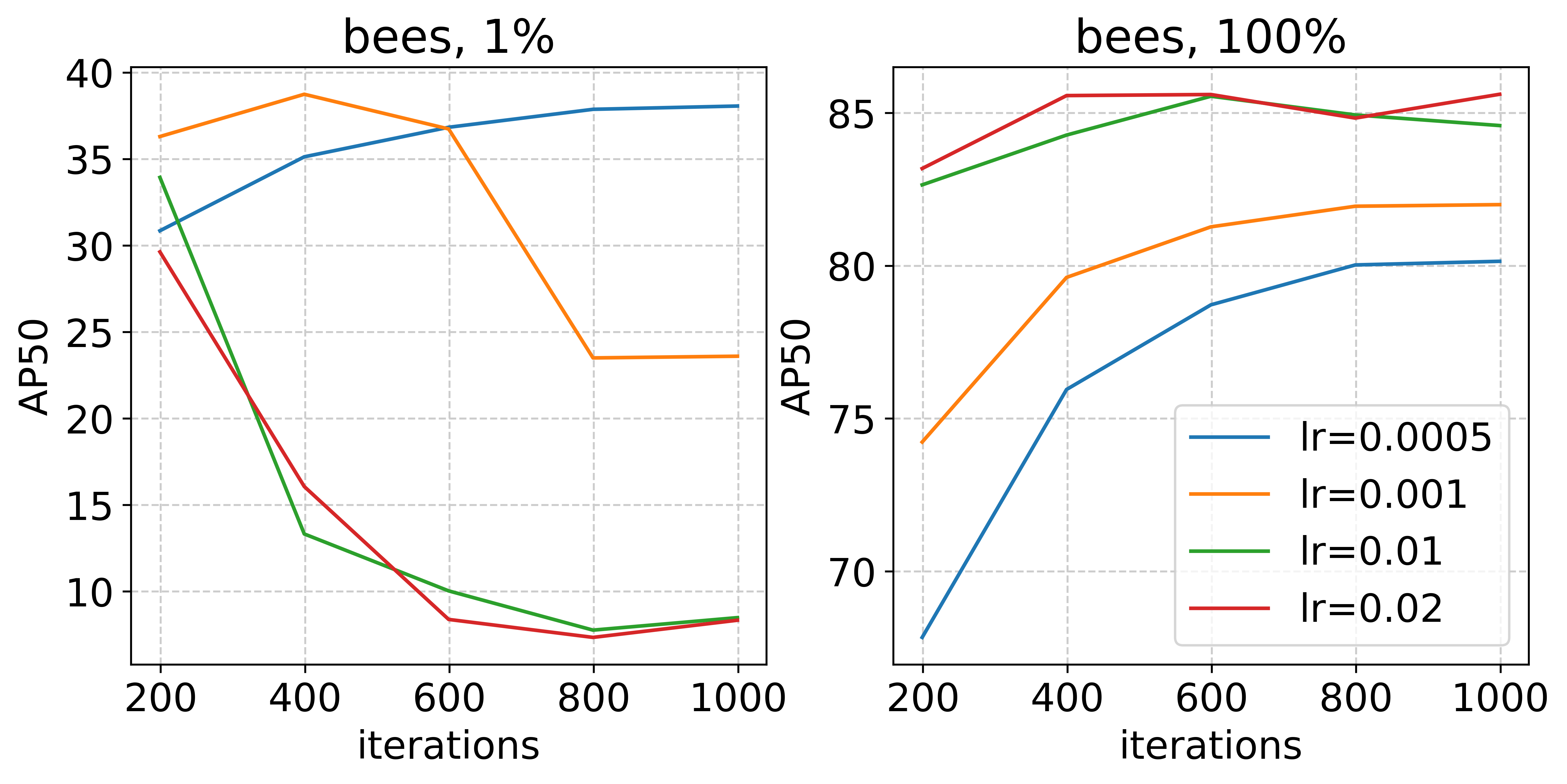}
\caption{Effect of learning rate on partial annotations versus full annotations on Minneapple and Bees.}
\label{F:lr_minneaple_bees}
\end{figure}

Figure~\ref{F:lr_minneaple_bees} shows the behavior of different learning rates for Faster R-CNN $\in$ \{0.0005, 0.001, 0.01, 0.2\} for Minneapple and Bees.
A higher learning rate for partial annotations is prone to over-fitting on missing annotations.
For example, learning rate of 0.01 or 0.02 that show good performance on the Bees dataset with full annotations severely overfit on 1\% partial annotations.
Based on this ablation study, we fix the learning rate for partial annotation scenarios at 0.0005 for all datasets. To get a more accurate measure of regret, for the performance with 100\% annotations, we show the numbers corresponding to the best learning rate, i.e., 0.02 in the case of Bees and 0.0005 in the case of Minneapple.

\end{document}